\documentclass{article} %
\usepackage{iclr2027_conference,times}

\usepackage{amsmath,amsfonts,bm}

\def\eqref#1{equation~\ref{#1}}
\def\Eqref#1{Equation~\ref{#1}}

\def\1{\bm{1}}

\DeclareMathAlphabet{\mathsfit}{\encodingdefault}{\sfdefault}{m}{sl}
\SetMathAlphabet{\mathsfit}{bold}{\encodingdefault}{\sfdefault}{bx}{n}

\usepackage{amsmath,amssymb}
\usepackage{graphicx}
\usepackage{booktabs}
\usepackage{multirow}
\usepackage{float}
\usepackage[section]{placeins}
\usepackage{algorithm}
\usepackage{algpseudocode}
\usepackage[table]{xcolor}
\usepackage{microtype}
\usepackage{url}
\definecolor{linkblue}{rgb}{0.21,0.49,0.74}
\usepackage[breaklinks,colorlinks,allcolors=linkblue]{hyperref}

\definecolor{lilac}{HTML}{EEE6F7}

\title{CARES: Coverage-Aware Replay\\Exemplar Selection for Continual Learning}

\author{\normalfont Arundhati S. Shanbhag$^{1,2}$, Brian B. Moser$^{2,3}$,\\
Federico Raue$^{2}$, Tobias C. Nauen$^{1,2}$, Andreas Dengel$^{1,2}$\\[0.5em]
{\normalfont\small $^{1}$ University of Kaiserslautern-Landau, Germany}\\
{\normalfont\small $^{2}$ German Research Center for Artificial Intelligence, Germany}\\
{\normalfont\small $^{3}$ Corresponding Author}\\
{\normalfont\small\texttt{first.last@dfki.de}}
}

\iclrfinalcopy

\begin{document}

\maketitle
\lhead{Preprint} %

\begin{abstract}
Replay-based Continual Learning mitigates catastrophic forgetting by keeping a buffer of past samples, making buffer composition critical under a fixed memory budget.
Recent coreset-based replay methods improve buffer quality using model-derived gradients, losses, or auxiliary optimization, but incur substantial computational overhead as memory is repeatedly updated. We propose CARES, a continual model-optimization-free approach for buffer construction that instead exploits the geometry of a frozen pretrained feature space using class-aware, density-weighted coverage. We evaluate CARES across balanced, imbalanced, and label-noisy streams under online class- and task-incremental settings. On Tiny-ImageNet, CARES is 2.3–8.8$\times$ faster than existing coreset-based selectors. Although CARES achieves this speedup using no training signal from the continual learner, it ranks among the state of the art across balanced, imbalanced, and label-noisy streams under online class- and task-incremental settings, outperforming the strongest baseline by up to 7.58 percentage points under class imbalance while also attaining higher average accuracy on Tiny-ImageNet. Code will be released upon publication.
\end{abstract}

\section{Introduction}
\label{sec:intro}

Continual learning (CL) studies how a model can acquire knowledge while being trained on a sequence of tasks without losing performance on previously learned tasks. This setting reflects realistic learning scenarios, where data does not arrive all at once but instead appears as a stream over time \citep{wang2024}. However, deep neural networks trained under this constraint suffer from \textit{catastrophic forgetting} \citep{McCloskey1989Catastrophic, Robins1995Catastrophic}, in which updates on newer tasks overwrite representations useful for older ones. This problem is particularly challenging in class-incremental learning, where the model must distinguish among all classes observed so far without receiving the task identity at inference time.

Methods for mitigating catastrophic forgetting are commonly grouped into regularization-, architecture-, and rehearsal-based approaches. \textit{Regularization} methods constrain model updates to preserve previously acquired knowledge \citep{Kirkpatrick_2017, aljundi2018memory, Chaudhry_2018, ritter2018, serra2018}. \textit{Architecture-based} approaches modify the network architecture through expansion or parameter isolation \citep{rusu2022progressive, schwarz2018progress, mallya2018, xu2018reinforced, yoon2020scalable, hung2019, sokar2021}. Finally, \textit{replay-based} methods maintain a buffer of samples from past tasks, which are replayed during training to address forgetting \citep{chaudhry2019agem, chaudhry2019tiny,buzzega2020dark, aljundi2019gss, aljundi2019ermir, lopezpaz2022gradient}.

Under a fixed memory budget, the effectiveness of rehearsal depends not only on the number of samples stored but also on which samples are retained in the buffer \citep{rebuffi2017icarl, yoon2022ocr, aljundi2019gss}. Methods such as experience replay \citep{chaudhry2019tiny} provide inexpensive buffer updates but do not select samples that accurately represent the previously seen data. Other works employ distillation losses or constrain updates in addition to replaying past samples \citep{buzzega2020dark, Chaudhry_2018, chaudhry2019agem}. Consequently, recent work has formulated buffer construction in rehearsal as a coreset selection problem \citep{yoon2022ocr, tiwari2022gcr, hao2023bilevel}. 

In a continual learning stream, buffer updates happen repeatedly and must keep pace with incoming data; selection is not a one-time cost, but one incurred at every task boundary or even every incoming batch \citep{wang2024}.
Gradient- and loss-based criteria \citep{aljundi2019gss, aljundi2019ermir, tiwari2022gcr, yoon2022ocr, hao2023bilevel} produce effective buffers, particularly under extreme memory limits, but they rely on additional computationally heavy optimization steps. Because this cost recurs at every update rather than once, it accumulates over the lifetime of the stream and can become a substantial fraction of total training time. This computational burden creates a practical tension: the very methods designed to make rehearsal more sample-efficient often introduce overhead that undermines the efficiency gains they seek to provide, particularly when buffer updates must keep pace with a continuous data stream \citep{tiwari2022gcr, aljundi2019ermir}.

These limitations raise a more fundamental question: \emph{must a continually updated model also provide the signal that determines what it should remember?} We investigate a complementary hypothesis: a replay buffer can instead be constructed by preserving the representative structure of the observed data in a stable feature space. Features from a frozen encoder provide a continual model-independent representation space in which examples can be compared throughout the stream. Within this space, geometric coverage encourages the buffer to represent distinct regions of the data distribution. However, pure coverage may overemphasize isolated candidates, whereas prioritizing only dense regions risks selecting redundant examples from dominant modes. Density-aware coverage provides a natural compromise between these two objectives.

SubZeroCore \citep{moser2025subzero} addresses this coverage-density tension in static coreset selection through a density-aware coverage objective. We build on this rather than classical geometric selectors such as herding \citep{welling2009, rebuffi2017icarl},  or k-Center Greedy \citep{sener2018active}, which primarily emphasize representativeness or coverage \citep{moser2026coreset}, because SubZeroCore incorporates density without requiring model-derived training signals. This makes it well suited to the repeated memory updates of online continual learning. However, it assumes simultaneous access to a fixed candidate pool, whereas continual learning requires a bounded buffer that must be updated as new task data arrives and the distribution of available candidates may become strongly imbalanced across classes. Directly applying a global geometric selector can therefore allocate disproportionate memory to classes with more candidates.

To address these challenges, we propose \textbf{CARES}, a continual model-optimization-free approach to replay-buffer construction for online continual learning. At each memory-update step, CARES combines the existing replay memory with a staging pool of newly observed examples and embeds the resulting candidate set using a frozen pretrained encoder. It then allocates the fixed replay budget across observed classes and applies density-aware coverage independently within each class before reconstructing the memory. In this way, CARES adapts training-free geometric coreset selection to a continually evolving replay buffer of fixed capacity. 

Crucially, memory construction in CARES is decoupled from optimization of the continual learner: selection requires no model-derived losses, gradients, or auxiliary model training. On Split-Tiny-ImageNet, this yields memory updates that are $2.3$–$8.8\times$ faster than existing coreset-based replay methods, while CARES attains higher final accuracy. This favorable accuracy–efficiency trade-off holds across balanced, imbalanced, and label-noisy online continual-learning settings, where CARES remains competitive with considerably more expensive selectors and is particularly effective under class imbalance.

\section{Related Work}
\label{sec:related-work}

\paragraph{Replay-based CL:} Experience Replay (ER) \citep{chaudhry2019tiny} trains jointly on current and stored examples. Beyond random storage, replay methods select or retrieve examples using
interference, uncertainty, gradients, or diversity. Dark Experience Replay (DER) and DER++ \citep{buzzega2020dark} apply distillation-style objectives, whereas others \citep{lopezpaz2022gradient, chaudhry2019agem} constrain gradient updates to preserve previous model behavior. GSS \citep{aljundi2019gss}, promotes gradient diversity among buffer samples, while MIR \citep{aljundi2019ermir} retrieves examples predicted to suffer the largest loss increase after an update. Collectively, these methods underscore that buffer composition, not just buffer size, drives rehearsal effectiveness.

\paragraph{Feature-Space Buffer Construction:} A separate line of work sidesteps learner-derived signals entirely, relying instead on a fixed feature extractor. iCaRL \citep{rebuffi2017icarl} uses the feature extractor of the continually trained network to select exemplars by herding.  REMIND \citep{hayes2020remind}, instead, stores compressed hidden representations from frozen early network layers for later replay, while Rainbow Memory \citep{bang2021rainbow} selects diverse exemplars based on classification uncertainty computed directly in feature space. Unlike these methods, CARES formulates buffer construction as an explicit coreset-selection objective over a frozen embedding space, combining class-aware allocation with density-aware coverage rather than uncertainty- or compression-driven heuristics.

\paragraph{Coreset-Selection for Replay:} A coreset is a small, representative subset of a larger dataset that approximates the full-data objective for a given learning task. Classical geometric constructions include clustering \citep{gonzalez1985clustering},
farthest-first traversal \citep{harpeled2004coresets}, and facility location \citep{cornuejols1977location}. Depending on the task, coresets may be selected to preserve geometric coverage, class balance, diversity, uncertainty, gradient information, or loss characteristics \citep{moser2026coreset}. In memory-constrained settings, coreset selection provides a principled way to decide which samples should be retained when only a limited number of examples can be stored. 

Applying coreset selection to buffer construction has gained traction in recent years. Recent CL methods optimize model-dependent proxies for buffer construction. GCR \citep{tiwari2022gcr} selects examples whose gradients approximate those of all data seen so far.  OCS \citep{yoon2022ocr} applies
representativeness, diversity, and affinity criteria during online training. On the other hand, BCSR \citep{hao2023bilevel} formulates selection as a bilevel problem over a learned sample distribution, while CSReL \citep{tong2025coreset} prioritizes examples through a
reducible-loss signal obtained with a holdout model. Across these works, a common thread emerges: effective rehearsal depends on formalizing which samples to keep as an explicit optimization problem rather than a heuristic rule, a principle that motivates the coreset-based approach taken in this work but, as we discuss next, this formalization typically comes at the cost of repeated continual model-dependent computations.

\section{CARES}
\label{sec:cares_methodology}

\begin{algorithm}[t] 
\caption{CARES Replay Memory Update} 
\label{alg:cares} 
    \begin{algorithmic}[1] 
        \Require Task stream $\mathcal{D}_{1:T}$, replay capacity $K$, staging capacity $P$, target coverage $\gamma$, frozen encoder $g_{\phi}$ 
        \State $\mathcal{M}_0 \gets \emptyset$ 
        \For{$t=1,\ldots,T$} 
            \State $\mathcal{P}_t \gets \emptyset$ 
            \For{each minibatch $\mathcal{B}_t \subset \mathcal{D}_t$}
                \State Sample a replay minibatch $\mathcal{B}^{\mathcal{M}}_t$ uniformly from $\mathcal{M}_{t-1}$, if non-empty 
                \State Update $\theta$ using cross-entropy on $\mathcal{B}_t \uplus \mathcal{B}^{\mathcal{M}}_t$ 
            \State Update $\mathcal{P}_t$ with examples in $\mathcal{B}_t$ using reservoir sampling
        \EndFor 
        \State $\mathcal{A}_t \gets \mathcal{M}_{t-1} \uplus \mathcal{P}_t$ 
        \If{$|\mathcal{A}_t| \leq K$} 
            \State $\mathcal{M}_t \gets \mathcal{A}_t$ 
        \Else
    \State $\mathcal{Z}_t \gets
        \{(z_i,y_i) : z_i =
        g_{\phi}(x_i)/\|g_{\phi}(x_i)\|_2,$
    \Statex \hspace{\algorithmicindent}
        $(x_i,y_i)\in\mathcal{A}_t\}$
    \State $\mathcal{I}_t \gets
        \Call{CARES-Select}{\mathcal{Z}_t,K,\gamma}$
    \State $\mathcal{M}_t \gets \mathcal{A}_t[\mathcal{I}_t]$
\EndIf
        \EndFor 
    \end{algorithmic} 
\end{algorithm}

\subsection{Setting and Problem Definition}
We consider continual learning over a sequence of tasks with labeled
data \(\mathcal{D}_1,\ldots,\mathcal{D}_T\).
In task-incremental learning (TIL), the task identity is available at
inference and predictions are restricted to the corresponding class
set. In class-incremental learning (CIL), task identity is unavailable
and the model must predict among all classes observed so far.

Replay-buffer construction maintains a replay memory
\(\mathcal{M}_t\) of capacity \(K\) after every task \(t\), drawn from
the data observed so far:
\begin{equation}
\label{eq:replay}
    \mathcal{M}_0=\emptyset,
    \qquad
    \mathcal{M}_t
    \subseteq
    \mathcal{D}_{1:t},
    \qquad
    |\mathcal{M}_t|\leq K,
    \qquad
    \mathcal{D}_{1:t}
    =
    \bigcup_{s=1}^{t}\mathcal{D}_s.
\end{equation}
During task \(t\), an incoming minibatch \(\mathcal{B}_t\) is combined
with a replay minibatch
\(\mathcal{B}^{\mathcal M}_t\) sampled from
\(\mathcal{M}_{t-1}\), the memory built after the previous task. CARES
changes only how \(\mathcal{M}_t\) is constructed; it does not modify
the continual model's training objective or inference rule.

Ideally, after observing task \(t<T\), replay-memory construction
would select the subset of all data seen so far whose replay best
preserves performance on those tasks once training continues:
\begin{equation}
\label{eq:rehearsal-coreset}
\mathcal{M}_t^{*}
\in
\operatorname*{arg\,min}_{
    \mathcal{M}\subseteq\mathcal{D}_{1:t},\;
    |\mathcal{M}|\leq K
}
\mathcal{L}_{1:t}
\!\left(
    \theta_{t+1}(\mathcal{M})
\right),
\end{equation}
where \(\theta_{t+1}(\mathcal{M})\) denotes the continual model obtained
by training on \(\mathcal{D}_{t+1}\) with \(\mathcal{M}\) as replay
memory, and \(\mathcal{L}_{1:t}\) measures performance over all tasks
observed so far. The memory built after task \(t\) can only influence
training from task \(t+1\) onward, so its value is realized in the
future rather than in the model \(\theta_t\) that has already been
trained.

This ideal objective is not directly available in online continual
learning. The next task \(\mathcal{D}_{t+1}\) has not been observed when
\(\mathcal{M}_t\) must be fixed, the complete history
\(\mathcal{D}_{1:t}\) cannot be retained under the memory constraint,
and evaluating candidate memories through \(\mathcal{L}_{1:t}\) would
require repeatedly optimizing the continual model. Practical methods
therefore rely on tractable selection proxies computed from the data
available at each memory-update point.

\subsection{Continual Replay-Memory Construction}
\label{subsec:buffer_update}
CARES adapts the density-aware geometric criterion of
SubZeroCore~\citep{moser2025subzero} to online continual replay under a fixed budget. SubZeroCore assumes simultaneous access to a fixed candidate set. Continual replay instead requires newly observed examples to be
collected without storing the complete stream, compared with previously
retained examples, and selected under one persistent memory
capacity. CARES addresses these requirements through candidate
staging, repeated memory reconstruction, and class-aware budget
allocation.

For notational simplicity, \(t\) indexes a protocol-defined
memory-update stage. During stage \(t\), the replay memory
\(\mathcal{M}_{t-1}\) remains available for training, while newly
observed examples are accumulated in a temporary staging pool
\(\mathcal{P}_t\) of capacity \(P\). Once more than \(P\) examples have
been observed, the staging pool is maintained using reservoir sampling.
This operation only bounds the current-stage candidate collection; it
does not determine the final replay memory.

At the memory-update point, CARES combines the previous replay memory
and the staging pool:
\begin{equation}
\label{eq:candidate_pool}
    \mathcal{A}_t
    =
    \mathcal{M}_{t-1}
    \uplus
    \mathcal{P}_t,
\end{equation}
where \(\uplus\) denotes the disjoint union of the two example
collections. They share no examples because
\(\mathcal{P}_t\subseteq\mathcal{D}_t\) and
\(\mathcal{M}_{t-1}\subseteq\mathcal{D}_{1:t-1}\).
Previously retained examples therefore compete directly with newly
observed candidates.

Each candidate is represented using a frozen pretrained encoder
\(g_{\phi}\):
\begin{equation}
\label{eq:frozen_embedding}
    z_i
    =
    \frac{g_{\phi}(x_i)}
         {\lVert g_{\phi}(x_i)\rVert_2},
    \qquad
    \mathcal{Z}_t
    =
    \{(z_i,y_i):(x_i,y_i)\in\mathcal{A}_t\}.
\end{equation}
Because \(g_\phi\) remains fixed throughout the stream, selection is
independent of changes in the continually trained model
\(f_{\theta}\). In particular, CARES uses no predictions, losses,
gradients, or parameters from \(f_{\theta}\), and trains no auxiliary
model for buffer construction.

If \(|\mathcal{A}_t|\leq K\), CARES retains all candidates. Otherwise,
it divides the replay capacity approximately equally across the classes
represented in \(\mathcal{A}_t\), applies density-aware coverage
selection independently within each class, and replaces the previous
memory with the candidates at the selected indices:
\begin{equation}
\label{eq:cares_memory_update}
    \mathcal{I}_t
    =
    \operatorname{CARES\textsc{-}Select}
    \left(
        \mathcal{Z}_t,K,\gamma
    \right),
    \qquad
    \mathcal{M}_t
    =
    \mathcal{A}_t[\mathcal{I}_t].
\end{equation}
In both cases
\(\mathcal{M}_t\subseteq\mathcal{A}_t\subseteq\mathcal{D}_{1:t}\) and
\(|\mathcal{M}_t|\leq K\), so the update preserves
~\Eqref{eq:replay} by induction over \(t\).
The staging pool is then cleared before the next update stage.
Algorithm~\ref{alg:cares} summarizes the complete procedure.

\subsection{Density-Aware Coverage Selection}
\label{subsec:subzerocore}

The density score and facility-location objective described below are
adopted from SubZeroCore~\citep{moser2025subzero}. CARES contributes
their continual application through bounded staging, repeated
competition between old and new candidates, and class-aware allocation
of the persistent replay capacity.

Consider a candidate collection with normalized embeddings
\(\{z_i\}_{i=1}^{N}\), from which \(m<N\) examples must be selected.
For a coverage target \(\gamma\in(0,1]\), the neighborhood order is
defined as
\begin{equation}
\label{eq:coverage}
\kappa
=
\min
\left\{
k\in\{1,\ldots,N-m\}:
1-
\prod_{q=1}^{k}
\frac{N-m-q}{N-q}
\geq\gamma
\right\}.
\end{equation}
The product equals \(\binom{N-1-k}{m}/\binom{N-1}{m}\), the probability
that a uniformly random selection of \(m\) of the other \(N-1\)
candidates contains none of the \(k\) nearest neighbors of a given
candidate. \Eqref{eq:coverage} therefore chooses the smallest
neighborhood through which a random selection of size \(m\) covers a
non-selected candidate with probability at least \(\gamma\). The set is
non-empty for every \(\gamma\leq 1\) because the product vanishes at
\(k=N-m\).
Let \(r_i\) be the Euclidean distance from \(z_i\) to its \(\kappa\)-th nearest neighbor among the other candidates in the normalized embedding space.
Given the empirical mean~\(\mu\) and standard deviation~\(\sigma\) of
these radii, each candidate receives the density score
\begin{equation}
\label{eq:density}
s_i
=
\exp
\left(
-\frac{(r_i-\mu)^2}
       {2\bar{\sigma}^{\,2}}
\right),
\qquad
\bar{\sigma}
=
\max(\sigma,\epsilon),
\end{equation}
where \(\epsilon>0\) prevents numerical instability. This score assigns
lower weights to candidates with unusually large radii, which are
relatively isolated, and to candidates with unusually small radii,
which lie in highly redundant regions.

Because the embeddings are normalized, the non-negative cosine
similarity used by CARES is
\begin{equation}
\label{eq:similarity}
\operatorname{sim}(z_i,z_j)
=
\max\left(0,z_i^{\top}z_j\right).
\end{equation}
For a selected index set
\(\mathcal{S}\subseteq\{1,\ldots,N\}\), the density-weighted
facility-location objective is
\begin{equation}
\label{eq:facility}
F(\mathcal{S})
=
\sum_{i=1}^{N}
\max_{j\in\mathcal{S}}
\left[
s_j\,
\operatorname{sim}(z_i,z_j)
\right].
\end{equation}
CARES approximately maximizes
\Eqref{eq:facility} under the cardinality constraint
\(|\mathcal{S}|\leq m\) using greedy marginal-gain selection.

\paragraph{Class-aware allocation.}

Directly optimizing \Eqref{eq:facility} over all candidates can
allow classes with more available examples to occupy a disproportionate
share of the replay memory. CARES therefore allocates the memory budget
across classes before applying geometric selection.

Let \(\mathcal{Y}_t\) be the labels represented in
\(\mathcal{A}_t\), let
\(J_t=|\mathcal{Y}_t|\), and let
\(N_c\) denote the number of candidates belonging to class \(c\).
Selection is only invoked when \(|\mathcal{A}_t|>K\), so the full
capacity \(K\) is distributed. Defining
\begin{equation}
\label{eq:class_quota}
    b
    =
    \left\lfloor
    \frac{K}{J_t}
    \right\rfloor,
    \qquad
    q
    =
    K\bmod J_t,
\end{equation}
each class receives an initial allocation \(b_c\in\{b,b+1\}\), where
the \(q\) classes with the largest \(N_c\) receive the additional slot,
so that \(\sum_{c}b_c=K\).
The effective target for class \(c\) is
\begin{equation}
\label{eq:effective_class_quota}
    m_c
    =
    \min(b_c,N_c).
\end{equation}

Classes with \(m_c=N_c\) are retained in full. For every other class,
CARES evaluates
\Eqref{eq:coverage}--\Eqref{eq:facility} on that class's
candidates with \(N=N_c\) and \(m=m_c\), and takes the union of the
resulting class-wise subsets. The resulting replay memory contains
\(\sum_{c}m_c\leq K\) examples, with equality unless some class has
fewer candidates than its allocation, and prevents classes with larger
candidate pools from automatically dominating the global geometric
objective.

\section{Experiments}
\label{sec:exps}

We evaluate CARES under two complementary online continual-learning protocols, measuring robustness to non-ideal streams, performance across memory capacities, and computational efficiency. Let \(a_{l,j}\) denote the test accuracy on task \(j\) after training on task \(l\). We report final average accuracy and final average forgetting~\citep{Chaudhry_2018},
\begin{equation}
\label{eq:metrics}
\begin{aligned}
    A_T
    &=
    \frac{1}{T}\sum_{j=1}^{T} a_{T,j},
    \\
    \mathrm{FGT}_T
    &=
    \frac{1}{T-1}\sum_{j=1}^{T-1}
    \max_{l\in\{j,\ldots,T-1\}}
    \left(a_{l,j}-a_{T,j}\right).
\end{aligned}
\end{equation}
Throughout, \(A_T\) is reported in percent and \(\mathrm{FGT}_T\) as a fraction, with task accuracies converted to fractions when computing forgetting. Under the definition above, forgetting can be negative if final accuracy exceeds the best earlier accuracy; negative values indicate improvement rather than forgetting.

\begin{table}[!t]
    \centering
    \caption{Results on Split-CIFAR-100. CARES matches the strongest baseline on the balanced stream and achieves the highest final average accuracy under imbalance and label noise settings.}
    \label{tab:cifar100_results_p1}

    \setlength{\tabcolsep}{5.5pt}
    \renewcommand{\arraystretch}{1.12}

    \resizebox{\textwidth}{!}{%
    \begin{tabular}{lcccccc}
        \toprule
        \multirow{2}{*}{\textbf{Method}}
        & \multicolumn{2}{c}{\textbf{Balanced}}
        & \multicolumn{2}{c}{\textbf{Imbalanced}}
        & \multicolumn{2}{c}{\textbf{Label Noise}} \\
        \cmidrule(lr){2-3}
        \cmidrule(lr){4-5}
        \cmidrule(lr){6-7}
        & $\mathrm{A_T}\uparrow$
        & $\mathrm{FGT_T}\downarrow$
        & $\mathrm{A_T}\uparrow$
        & $\mathrm{FGT_T}\downarrow$
        & $\mathrm{A_T}\uparrow$
        & $\mathrm{FGT_T}\downarrow$ \\
        \midrule

        K-means Embedding
        & $59.77{\pm}0.24$
        & $0.061{\pm}0.001$
        & $43.91{\pm}0.15$
        & $0.044{\pm}0.001$
        & $57.92{\pm}1.25$
        & $0.091{\pm}0.016$ \\

        Uniform
        & $58.99{\pm}0.54$
        & $0.074{\pm}0.004$
        & $44.73{\pm}0.11$
        & $0.033{\pm}0.007$
        & $58.76{\pm}1.07$
        & $0.087{\pm}0.006$ \\

        iCaRL
        & $60.74{\pm}0.09$
        & $\mathbf{0.044{\pm}0.026}$
        & $44.25{\pm}2.04$
        & $0.042{\pm}0.019$
        & $59.70{\pm}0.70$
        & $0.071{\pm}0.010$ \\

        Greedy Coreset
        & $59.39{\pm}0.16$
        & $0.066{\pm}0.017$
        & $43.80{\pm}0.01$
        & $0.039{\pm}0.007$
        & $58.22{\pm}0.16$
        & $0.066{\pm}0.001$ \\

        GCR
        & $58.73{\pm}0.43$
        & $0.073{\pm}0.013$
        & $44.48{\pm}0.05$
        & $0.035{\pm}0.005$
        & $58.72{\pm}0.63$
        & $0.081{\pm}0.005$ \\

        OCS
        & $52.57{\pm}0.37$
        & $0.088{\pm}0.001$
        & $46.54{\pm}0.34$
        & $\mathbf{0.022{\pm}0.003}$
        & $51.77{\pm}0.81$
        & $0.103{\pm}0.007$ \\

        BCSR
        & $\mathbf{61.60{\pm}0.14}$
        & $0.051{\pm}0.015$
        & $47.30{\pm}0.57$
        & $\mathbf{0.022{\pm}0.005}$
        & $60.70{\pm}0.08$
        & $\mathbf{0.059{\pm}0.013}$ \\

        \midrule
        \rowcolor{lilac}
        \textbf{CARES (Ours)}
        & $61.52{\pm}1.48$
        & $0.111{\pm}0.074$
        & $\mathbf{54.88{\pm}0.33}$
        & $0.082{\pm}0.083$
        & $\mathbf{62.12 {\pm} 1.14}$
        & $0.110{\pm}0.013$ \\

        \bottomrule
    \end{tabular}%
    }
\end{table}

\subsection{Experimental Setup}

\paragraph{Protocol 1:}
Following BCSR~\citep{hao2023bilevel}, we evaluate online TIL under
balanced, long-tailed imbalanced, and noisy-label streams.
Split CIFAR-100~\citep{krizhevsky2009learning} and Split Tiny-ImageNet~\citep{Le2015TinyIV}
divide the CIFAR-100 \citep{krizhevsky2009learning} and TinyImageNet \citep{pouransari2014tinyimagenet} datasets into 20 tasks of 5 and 10 disjoint classes, respectively, with
replay capacities \(K=100\) and \(K=200\). CARES replaces only the
replay-memory selection procedure while retaining the reference
training and evaluation protocol.

\paragraph{Protocol 2:}
Following GCR~\citep{tiwari2022gcr}, we evaluate online CIL on
Split-CIFAR-10, Split-CIFAR-100, and Split-TinyImageNet. Split-CIFAR-10 and Split-CIFAR-100 are split into 5 tasks of 2 and 20 non-overlapping classes respectively whereas Split-TinyImageNet is split into 10 tasks of 20 classes. The model receives a single pass over streams with
blurred task boundaries and is evaluated at
\(K\in\{200,500,2000\}\), without task identity at inference.

\paragraph{Baselines.}
The comparison set is protocol-specific. For each protocol, we retain
the baselines, data splits, learner architecture, replay objective, and
training configuration of the corresponding reference setting. We
therefore interpret differences only within a shared protocol and do
not compare absolute accuracies across tables. 

\paragraph{Implementation and reproducibility details.}
Unless otherwise stated, the continual model uses a ResNet-18 backbone,
while CARES constructs the replay memory using $\ell_2$-normalized
features from a frozen ImageNet-pretrained Inception-V3 encoder. Replay
minibatches match the current-training minibatch size (10/20 for
Split CIFAR-100/Tiny-ImageNet in Protocol~1 and 32 in Protocol~2).
Runtime measurements are obtained on a single NVIDIA
RTX A6000 GPU with four CPU cores. Full dataset-specific
hyperparameters are listed in Appendix~\ref{sec:supp_implementation}.

\subsection{Robustness to Imbalance and Label Noise}
\begin{table}[!htbp]
    \centering
    \caption{Experimental results on Split-Tiny-ImageNet. CARES achieves the highest average
accuracy ($A_T$) across all three stream conditions, with particularly large gains under label noise.}
    \label{tab:tiny_results_p1}

    \small
    \setlength{\tabcolsep}{5.5pt}
    \renewcommand{\arraystretch}{1.08}

    \resizebox{\textwidth}{!}{%
    \begin{tabular}{lcccccc}
        \toprule
        \multirow{2}{*}{\textbf{Method}}
        & \multicolumn{2}{c}{\textbf{Balanced}}
        & \multicolumn{2}{c}{\textbf{Imbalanced}}
        & \multicolumn{2}{c}{\textbf{Label Noise}} \\
        \cmidrule(lr){2-3}
        \cmidrule(lr){4-5}
        \cmidrule(lr){6-7}
        & $\mathrm{A_T}\uparrow$
        & $\mathrm{FGT_T}\downarrow$
        & $\mathrm{A_T}\uparrow$
        & $\mathrm{FGT_T}\downarrow$
        & $\mathrm{A_T}\uparrow$
        & $\mathrm{FGT_T}\downarrow$ \\
        \midrule

        K-means Embedding
        & $41.48{\pm}1.21$
        & $0.129{\pm}0.007$
        & $36.29{\pm}0.23$
        & $0.085{\pm}0.003$
        & $36.01{\pm}1.51$
        & $0.083{\pm}0.005$ \\

        Uniform
        & $42.11{\pm}0.52$
        & $0.129{\pm}0.002$
        & $37.07{\pm}0.53$
        & $0.083{\pm}0.009$
        & $37.14{\pm}1.05$
        & $0.099{\pm}0.003$ \\

        iCaRL
        & $43.84{\pm}0.09$
        & $0.114{\pm}0.004$
        & $37.65{\pm}0.84$
        & $0.058{\pm}0.003$
        & $38.52{\pm}0.25$
        & $0.063{\pm}0.006$ \\

        Greedy Coreset
        & $41.02{\pm}0.33$
        & $0.119{\pm}0.017$
        & $33.43{\pm}0.86$
        & $0.103{\pm}0.002$
        & $36.37{\pm}0.16$
        & $0.079{\pm}0.006$ \\

        GCR
        & $41.45{\pm}0.35$
        & $0.125{\pm}0.008$
        & $36.08{\pm}0.62$
        & $0.072{\pm}0.017$
        & $37.46{\pm}0.40$
        & $0.115{\pm}0.011$ \\

        OCS
        & $41.29{\pm}0.09$
        & $0.112{\pm}0.001$
        & $35.09{\pm}0.60$
        & $\mathbf{0.036{\pm}0.011}$
        & $35.36{\pm}0.94$
        & $0.061{\pm}0.005$ \\

        BCSR
        & $44.13{\pm}0.33$
        & $0.106{\pm}0.001$
        & $38.59{\pm}0.11$
        & $0.070{\pm}0.004$
        & $40.72{\pm}0.56$
        & $\mathbf{0.055{\pm}0.006}$ \\

        \midrule
        \rowcolor{lilac}
        \textbf{CARES (Ours)}
        &$\mathbf{49.49{\pm}0.41}$ & $\mathbf{0.099{\pm}0.015}$
        &$\mathbf{42.94{\pm}0.40}$ & $0.104{\pm}0.011$
        &$\mathbf{49.37{\pm}0.95}$ & $0.078{\pm}0.021$ \\

        \bottomrule
    \end{tabular}%
    }
\end{table}

\label{sec:robustness}

One of the main advantages of CARES is particularly pronounced under non-ideal streams, as demonstrated in Tables \ref{tab:tiny_results_p1} and \ref{tab:cifar100_results_p1}. On Split CIFAR-100, CARES reaches \(54.88\%\) under class imbalance, outperforming the strongest
baseline, BCSR, by \(7.58\) percentage points, while remaining competitive on the
balanced stream. On Tiny-ImageNet, CARES achieves the highest accuracy
under all three conditions, improving over BCSR by \(5.36\), \(4.35\),
and \(8.65\) points on balanced, imbalanced, and label-noisy streams,
respectively. These results indicate that learner-independent,
class-aware geometric selection remains effective when the stream
distribution is skewed, or labels are corrupted.

CARES does not uniformly minimize $\mathrm{FGT}_T$, but this does not
contradict its stronger end-of-stream performance. While $A_T$ measures
the absolute accuracy retained after the full stream, $\mathrm{FGT}_T$
measures the decline of each task relative to its best historical
accuracy. Consequently, a method can exhibit a larger peak-to-final
drop while still retaining higher absolute accuracy at the end of
training. We therefore view CARES's primary advantage in Protocol~1 as
improved final retained performance, with $A_T$ and $\mathrm{FGT}_T$
capturing complementary aspects of continual-learning behavior.

\subsection{Performance Across Memory Capacities}
\label{sec:online_results}

Table~\ref{tab:cil_avg_accuracy} evaluates CARES under online CIL at different replay capacities. At \(K=200\), CARES is competitive on Split-CIFAR-10 and Split-CIFAR-100 and
achieves the highest accuracy on Split-TinyImageNet, reaching \(6.60\%\)
compared with \(6.19\%\) for GCR. At \(K=2000\), CARES achieves the
highest mean accuracy on Split-CIFAR-10 and Split-CIFAR-100, reaching
\(54.88\%\) and \(25.17\%\), respectively. These results indicate that
CARES benefits strongly from additional replay capacity while remaining
competitive under tighter memory constraints. Results for the
intermediate \(K=500\) setting are provided in
Appendix~\ref{sec:supp_robustness} (Table~\ref{tab:cil_avg_accuracy_full}).

\subsection{Runtime Efficiency}
\begin{figure}[!htbp]
    \centering
    \includegraphics[width=0.6\linewidth]{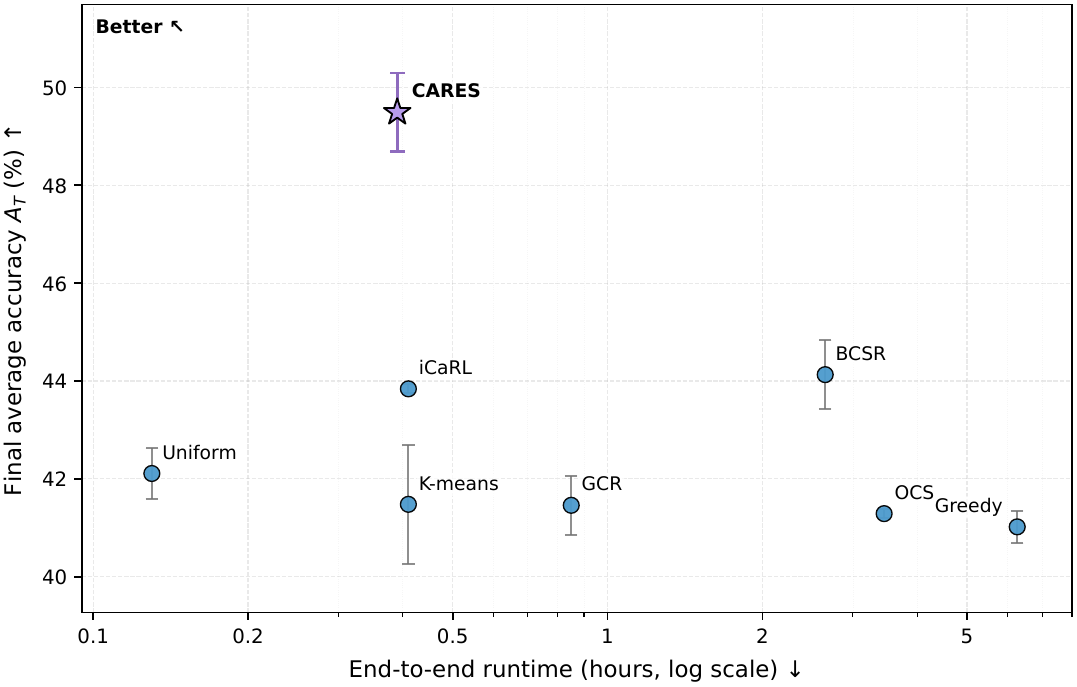}
    \caption{Accuracy--runtime trade-off on Split-Tiny-ImageNet under the balanced setting in Protocol 1 at $K=200$.
    CARES achieves substantially higher final accuracy at a runtime comparable
    to lightweight selection methods.}
    \label{fig:runtime_accuracy}
\end{figure}

\label{sec:runtime}
Figure~\ref{fig:runtime_accuracy} compares accuracy and end-to-end runtime on Split-Tiny-ImageNet under
Protocol~1. CARES completes the experiment in \(0.39\) hours, comparable
to K-means Embedding and iCaRL (\(0.41\) hours), while being
\(2.3\times\), \(6.8\times\), and \(8.8\times\) faster than GCR, BCSR,
and OCS, respectively. Importantly, CARES also achieves higher final
accuracy than these methods, yielding a favorable accuracy--efficiency
trade-off without learner-derived selection signals.

\begin{table}[t]
\centering
\caption{Online CIL average accuracy (15 seeds) under small ($K=200$) and large ($K=2000$) replay budgets. CARES is competitive
under tight memory and becomes increasingly effective as capacity grows,
achieving the highest mean accuracy on Split-CIFAR-10 and
Split-CIFAR-100 at \(K=2000\).}
\label{tab:cil_avg_accuracy}
\scriptsize
\setlength{\tabcolsep}{6pt}
\resizebox{\textwidth}{!}{%
\begin{tabular}{lcccccc}
\toprule
\multirow{2}{*}{\textbf{Method}} 
& \multicolumn{2}{c}{\textbf{Split-CIFAR-10}} 
& \multicolumn{2}{c}{\textbf{Split-CIFAR-100}} 
& \multicolumn{2}{c}{\textbf{Split-TinyImageNet}} \\
\cmidrule(lr){2-3} 
\cmidrule(lr){4-5} 
\cmidrule(lr){6-7}
& $K=200$ & $K=2000$
& $K=200$ & $K=2000$
& $K=200$ & $K=2000$ \\
\midrule

ER 
& $32.63{\pm}5.11$ & $49.31{\pm}7.45$
& $11.72{\pm}1.25$ & $20.78{\pm}1.48$
& $4.36{\pm}1.02$ & $11.30{\pm}1.45$ \\

GEM 
& $16.81{\pm}1.17$ & $20.35{\pm}2.44$
& $10.42{\pm}1.66$ & $12.38{\pm}3.81$
& $5.15{\pm}0.71$ & $9.56{\pm}2.63$ \\

GSS 
& $33.83{\pm}3.95$ & $41.70{\pm}4.56$
& $12.31{\pm}0.57$ & $16.24{\pm}0.80$
& $5.82{\pm}0.38$ & $8.44{\pm}0.48$ \\

iCaRL 
& $38.66{\pm}2.03$ & $39.56{\pm}1.53$
& $11.23{\pm}0.31$ & $12.28{\pm}0.39$
& $4.39{\pm}0.19$ & $6.66{\pm}0.30$ \\

MIR 
& $21.50{\pm}0.70$ & $46.40{\pm}1.81$
& $10.30{\pm}0.30$ & $16.02{\pm}0.50$
& $4.90{\pm}0.17$ & $7.05{\pm}0.30$ \\

DER 
& $38.26{\pm}5.08$ & $49.90{\pm}7.55$
& $10.68{\pm}2.03$ & $24.83{\pm}1.58$
& $4.81{\pm}0.57$ & $\mathbf{15.97{\pm}0.95}$ \\

GCR 
& $\mathbf{42.26{\pm}7.60}$ & $52.30{\pm}7.01$
& $\mathbf{14.88{\pm}0.94}$ & $22.47{\pm}2.01$
& $6.19{\pm}0.36$ & $11.83{\pm}1.26$ \\

\midrule
\rowcolor{lilac}
\textbf{CARES (Ours)}
& $40.26{\pm}4.67$ & $\mathbf{54.88{\pm}6.34}$
& $14.28{\pm}0.84$ & $\mathbf{25.17{\pm}1.39}$
& $\mathbf{6.60{\pm}0.59}$ & $15.67{\pm}0.85$ \\

\bottomrule
\end{tabular}%
}
\end{table}

\subsection{Sensitivity Analysis}
\label{sec:sensitivity}

Table~\ref{tab:hyperparam_sensitivity} studies sensitivity to the
coverage parameter \(\gamma\) and staging-pool capacity \(P\) on balanced
Split-CIFAR-100. Performance remains relatively stable across a broad
range of coverage values, while increasing the staging pool beyond a
moderate multiple of the replay capacity provides no consistent gain.
This indicates that CARES does not require precise tuning or an
excessively large temporary candidate pool.

\begin{table}[h]
    \centering
    \small
    \setlength{\tabcolsep}{4.5pt}
    \caption{Hyperparameter sensitivity of CARES on balanced Split CIFAR-100. Final task-incremental accuracy (\%) when varying the coverage parameter $\gamma$ and the staging-pool size $P$, expressed as a multiple of the replay capacity $K$. All other hyperparameters are held fixed.}
    \label{tab:hyperparam_sensitivity}
    \begin{tabular}{l*{11}{c}}
        \toprule
        $\gamma$ & 0.1 & 0.2 & 0.3 & 0.4 & 0.5 & 0.6 & 0.7 & 0.8 & 0.9 & 1.0 & \\
        $A_T$ & 59.8 & 59.0 & 61.3 & 59.5 & 60.0 & 58.7 & 60.6 & 59.8 & 60.8 & 60.1 & \\
        \midrule
        $P/K$ & 1 & 2 & 3 & 4 & 5 & 6 & 7 & 8 & 9 & 10 & 20 \\
        $A_T$ & 59.8 & 60.2 & 59.4 & 60.7 & 58.7 & 59.7 & 57.3 & 58.5 & 58.3 & 58.8 & 60.7 \\
        \bottomrule
    \end{tabular}
\end{table}

Table~\ref{tab:encoder_sensitivity} further evaluates the frozen
representation used for selection. CARES exhibits similar performance
across different pretrained encoders, indicating that its effectiveness
is not tied to a particular feature extractor.

\begin{table}[h]
    \centering
    \small
    \caption{Sensitivity to the frozen encoder. Final class-incremental accuracy (\%) of CARES on Split-CIFAR-100 under Protocol~2 with different pretrained encoders used for selection. CARES remains consistent across encoders, indicating limited sensitivity to the choice of representation used for coreset selection.}
    \label{tab:encoder_sensitivity}
    \begin{tabular}{lccc}
        \toprule
        \textbf{Selection encoder} & $K=200$ & $K=500$ & $K=2000$ \\
        \midrule
        InceptionV3    & 15.0 & 19.1 & 23.5 \\
        ResNet-50      & 14.4 & 19.6 & 23.7 \\
        CLIP ViT-B/32  & 14.3 & 19.1 & 23.3 \\
        \bottomrule
    \end{tabular}
\end{table}

\subsection{Ablation Studies}
\label{sec:ablation}

Table~\ref{tab:cares_ablation} isolates the contributions of
class-aware allocation, geometric coverage, and density weighting.
CB-Random preserves class-balanced allocation but selects samples
randomly, Global Facility applies coverage globally, Global SubZeroCore
adds density weighting to global coverage, and Class-wise Facility
performs geometric selection independently within each class.

The largest effect appears under class imbalance: class-wise selection
substantially outperforms both global variants, showing that explicit
allocation of replay capacity across classes is critical when candidate
frequencies are skewed. Density weighting provides complementary gains
on balanced and label-noisy streams, although its benefit is not uniform
across all conditions. Overall, the ablation shows that class-aware
allocation is the key continual-learning adaptation, with density-aware
coverage providing an additional refinement.

\begin{table}[t]
\centering
\small

\caption{
Component ablation of CARES on Split-CIFAR-100. Class-aware allocation provides
the dominant gain under class imbalance, substantially outperforming
global geometric selection, while density weighting provides additional
gains on balanced and label-noisy streams.
}

\label{tab:cares_ablation}

\setlength{\tabcolsep}{5pt}

\resizebox{\textwidth}{!}{%

\begin{tabular}{lcccccc}

\toprule

\textbf{Variant}
&
\textbf{Class-aware}
&
\textbf{Coverage}
&
\textbf{Density}
&
\textbf{Balanced}
&
\textbf{Imbalanced}
&
\textbf{Label Noise}
\\

\midrule

Uniform
&
$\times$
&
$\times$
&
$\times$
&
$58.99{\pm}0.54$
&
$44.73{\pm}0.11$
&
$58.76{\pm}1.07$
\\

CB-Random
&
$\checkmark$
&
$\times$
&
$\times$
&
$59.50{\pm}1.63$
&
$52.89{\pm}0.41$
&
$59.37{\pm}0.72$
\\

Global SubZeroCore
&
$\times$
&
$\checkmark$
&
$\checkmark$
&
$59.77{\pm}1.94$
&
$46.72{\pm}1.24$
&
$60.71{\pm}1.50$
\\
Global Facility
&
$\times$
&
$\checkmark$
&
$\times$
&
$59.83{\pm}1.50$
&
$46.29{\pm}2.07$
&
$61.36{\pm}1.09$
\\

Class-wise Facility
&
$\checkmark$
&
$\checkmark$
&
$\times$
&
$60.04{\pm}1.61$
&
$\mathbf{55.12{\pm}0.37}$
&
$61.37{\pm}1.41$
\\

\midrule

\rowcolor{lilac}
\textbf{CARES}
&
$\checkmark$
&
$\checkmark$
&
$\checkmark$
&
$\mathbf{61.52{\pm}1.48}$
&
$54.88{\pm}0.33$
&
$\mathbf{62.12{\pm}1.14}$
\\

\bottomrule

\end{tabular}
}

\end{table}

\section{Limitations and Future Work}
\label{sec:limitations}

CARES is less consistently advantageous under very small replay
budgets. When only a few exemplars can be retained, learner-dependent
selection signals can provide more targeted sample choices, whereas
CARES becomes increasingly competitive as additional replay capacity
allows the selected set to represent the feature-space geometry more
fully. Developing lightweight hybrid criteria that combine geometric
coverage with inexpensive uncertainty or learner-derived information
may improve this small-budget regime without sacrificing the efficiency
of CARES.

CARES also assumes access to a suitable frozen pretrained encoder.
Our sensitivity analysis indicates that performance is relatively
stable across the evaluated representations, but selection quality
ultimately depends on the structure induced by the pretrained feature
space. The staging pool introduces temporary memory beyond the permanent
replay budget; however, our sensitivity analysis shows no consistent
benefit from excessively large candidate pools, suggesting that moderate
staging capacities are sufficient in the evaluated setting.

\section{Conclusion}
\label{sec:conclusion}

We presented CARES, a learner-optimization-free approach to replay-buffer
construction that selects exemplars using class-aware, density-weighted
coverage in a frozen pretrained feature space. By decoupling memory
selection from the continually trained model, CARES requires no continual
learner-derived losses, gradients, backward passes, or auxiliary model
training during buffer construction.

Across online task- and class-incremental settings, CARES achieves
competitive or superior final accuracy while substantially reducing the
cost of memory selection compared with learner-dependent coreset methods.
Its advantage is particularly pronounced under class imbalance, and its
relative performance improves as additional replay capacity becomes
available. Our ablations further show that class-aware allocation is
critical for geometric selection under imbalanced streams, while density
weighting provides complementary gains in balanced and label-noisy
settings.

These results demonstrate that effective replay-buffer construction need
not depend on signals from the continual learner itself. Stable pretrained
representation geometry provides a simple and computationally efficient
alternative, particularly when memory selection must be performed
repeatedly throughout an online data stream.

\subsection*{AI use statement}
In this work, we used generative AI tools to assist with different aspects
of the research and manuscript preparation. Specifically, it
was used to check and refine the mathematical formulation of the method
in Section~\ref{sec:cares_methodology}, to provide feedback on the design
and presentation of experiments, and to assist in refining and debugging
parts of the implementation. Generative AI was also used to edit the
manuscript for clarity and readability and to assist with formatting the
paper and its references for this template.

The authors reviewed all AI-assisted suggestions before being
incorporated. The mathematical formulation was checked against the
reference implementation, and all experimental procedures, code changes,
results, and interpretations were verified by the authors. We take full
responsibility for the final content of this work, including any text,
code, claims, or other artifacts produced with the assistance of
generative AI.

\subsection*{Ethics statement}
This work uses only publicly available image-classification benchmarks
(Split CIFAR-10, Split CIFAR-100, and Split Tiny-ImageNet) and involves no
human subjects and no personally identifiable data. The proposed method
changes which examples a continual learner replays and raises no ethical
concerns beyond those of the underlying datasets and of continual-learning
applications in general.

\subsection*{Reproducibility statement}
Algorithm~\ref{alg:cares} and Appendix~\ref{sec:algo} give the complete
memory-update procedure, and Section~\ref{sec:cares_methodology} states
every equation the selector evaluates. Appendix~\ref{sec:supp_protocols}
describes both evaluation protocols, including dataset splits, replay
capacities, and the class-imbalance and label-noise procedures.
Appendix~\ref{sec:supp_implementation} lists all hyperparameters
(Tables~\ref{tab:protocol1_hparams} and~\ref{tab:protocol2_hparams}), and
the evaluation metrics are defined in ~\eqref{eq:metrics}. Code will
be released upon publication.

\bibliography{main}
\bibliographystyle{iclr2027_conference}

\clearpage
\appendix
\section*{\centering Supplementary Material}

\section{CARES Algorithm}
\label{sec:algo}

Algorithm~\ref{alg:cares_full} provides the complete memory-update
procedure used by CARES. During task \(t\), incoming examples are
maintained in a bounded staging pool \(\mathcal{P}_t\) using reservoir
sampling. At the memory-update point, this pool is combined with the
previous replay memory \(\mathcal{M}_{t-1}\). If the resulting
candidate set exceeds the replay-memory budget \(K\), CARES embeds the
candidates using a frozen feature extractor and applies the
class-aware, density-weighted coverage criterion described in the
main paper.

\begin{algorithm}[h]
\caption{CARES: Coverage-Aware Replay Exemplar Selection}
\label{alg:cares_full}
\begin{algorithmic}[1]

\Require Task sequence $\{\mathcal{D}_t\}_{t=1}^{T}$,
         replay capacity $K$,
         staging-pool capacity $P$,
         target coverage $\gamma$,
         frozen feature extractor $g_{\phi}$

\Ensure Continually updated learner $f_{\theta}$ and replay memory
$\mathcal{M}_T$

\State Initialize replay memory $\mathcal{M}_0 \gets \emptyset$

\For{$t = 1,\ldots,T$}

    \State Initialize staging pool
    $\mathcal{P}_t \gets \emptyset$

    \For{each minibatch
    $\mathcal{B}_t \subset \mathcal{D}_t$}

        \Comment{Standard rehearsal training}

        \If{$\mathcal{M}_{t-1} \neq \emptyset$}

            \State Sample replay minibatch
            $\mathcal{B}_{\mathcal{M}}
            \sim \mathcal{M}_{t-1}$

            \State
            $\mathcal{B}
            \gets
            \mathcal{B}_t
            \uplus
            \mathcal{B}_{\mathcal{M}}$

        \Else

            \State
            $\mathcal{B}
            \gets
            \mathcal{B}_t$

        \EndIf

        \State Update $\theta$ using cross-entropy loss on
        $\mathcal{B}$

        \Comment{Bounded current-task staging}

        \For{each $(x,y)\in\mathcal{B}_t$}

            \State
            $\mathcal{P}_t
            \gets
            \Call{ReservoirUpdate}
            {\mathcal{P}_t,(x,y),P}$

        \EndFor

    \EndFor

    \Comment{Memory reconstruction}

    \State
    $\mathcal{A}_t
    \gets
    \mathcal{M}_{t-1}
    \uplus
    \mathcal{P}_t$

    \If{$|\mathcal{A}_t|\leq K$}

        \State
        $\mathcal{M}_t
        \gets
        \mathcal{A}_t$

    \Else

        \State Compute normalized frozen embeddings
        $z_i\gets g_{\phi}(x_i)/\lVert g_{\phi}(x_i)\rVert_2$
        for all $(x_i,y_i)\in\mathcal{A}_t$

        \State
        $\mathcal{I}_t
        \gets
        \Call{CARES-Select}
        {\{(z_i,y_i)\}_{i=1}^{|\mathcal{A}_t|},K,\gamma}$

        \Comment{Selection criterion defined in the main paper}

        \State
        $\mathcal{M}_t
        \gets
        \mathcal{A}_t[\mathcal{I}_t]$

    \EndIf

\EndFor

\end{algorithmic}
\end{algorithm}

\section{Experimental Protocol Details}
\label{sec:supp_protocols}
\subsection{Protocol 1:}
\label{sec:supp_protocol1}

Protocol~1 follows the experimental setting of BCSR \citep{hao2023bilevel}.
Split CIFAR-100 is divided into 20 tasks containing 5 disjoint
classes each, while Tiny-ImageNet is divided into 20 tasks containing 10 disjoint classes each. The total replay-memory capacity is \(K=100\) for Split CIFAR-100 and \(K=200\) for Tiny-ImageNet. We follow the exact imbalance and noise procedures as \citet{hao2023bilevel}.  Incoming stream batches contain 50 samples on Split CIFAR-100 and 100 samples on Tiny-ImageNet. 

Task identity is available at inference, corresponding to
task-incremental learning (TIL). For the robustness experiments, we
follow the same class-imbalance and label-noise generation procedures as BCSR \citep{hao2023bilevel}. CARES changes only the replay-memory selection procedure; the backbone, continual-model optimization, data stream, and evaluation protocol are otherwise kept consistent with the reference setting. Hyperparameters specific to BCSR's bilevel
optimization are used only for that baseline and are not applicable to CARES. Results are aggregated over three independent runs.

\subsection{Protocol 2}
\label{sec:supp_protocol2}

Protocol~2 follows the online-streaming setting of GCR \citep{tiwari2022gcr}.
Split-CIFAR-10 is divided into 5 tasks of 2 classes,
Split-CIFAR-100 into 5 tasks of 20 classes, and
Split-TinyImageNet into ten tasks of 20 classes.

Data are presented in streams of 6,250 samples with blurred task boundaries, and the learner receives a single pass over each stream. Evaluation is class-incremental: task identity is unavailable at inference, and the model predicts over all classes observed so far. We use a ResNet-18 learner with minibatches of 32 current examples
and 32 replay examples, and evaluate replay-memory capacities
\(K\in\{200,500,2000\}\). Following \citet{tiwari2022gcr}, methods that require boundary information may receive it during training, while evaluation remains class-incremental. Results are aggregated over 15 independent runs.

\section{Additional Results}
\label{sec:supp_robustness}

\begin{table}[h]
    \centering
    \small

    \caption{
    Results on Split-CIFAR-100 under balanced, imbalanced, and
    label-noise settings. Higher $\mathrm{A_T}$ is better, while
    lower $\mathrm{FGT_T}$ is better. Results are reported as mean
    $\pm$ standard deviation. Best results are shown in
    \textbf{bold}.
    }

    \label{tab:cifar100_full_results}

    \setlength{\tabcolsep}{5.5pt}
    \renewcommand{\arraystretch}{1.12}

    \resizebox{\textwidth}{!}{%

    \begin{tabular}{lcccccc}

        \toprule

        \multirow{2}{*}{\textbf{Method}}
        & \multicolumn{2}{c}{\textbf{Balanced}}
        & \multicolumn{2}{c}{\textbf{Imbalanced}}
        & \multicolumn{2}{c}{\textbf{Label Noise}} \\

        \cmidrule(lr){2-3}
        \cmidrule(lr){4-5}
        \cmidrule(lr){6-7}

        &
        $\mathrm{A_T}\uparrow$
        &
        $\mathrm{FGT_T}\downarrow$
        &
        $\mathrm{A_T}\uparrow$
        &
        $\mathrm{FGT_T}\downarrow$
        &
        $\mathrm{A_T}\uparrow$
        &
        $\mathrm{FGT_T}\downarrow$
        \\

        \midrule

        K-means Features
        & $57.82{\pm}0.69$
        & $0.070{\pm}0.003$
        & $45.44{\pm}0.76$
        & $0.037{\pm}0.002$
        & $57.38{\pm}1.26$
        & $0.098{\pm}0.003$
        \\

        K-means Embedding
        & $59.77{\pm}0.24$
        & $0.061{\pm}0.001$
        & $43.91{\pm}0.15$
        & $0.044{\pm}0.001$
        & $57.92{\pm}1.25$
        & $0.091{\pm}0.016$
        \\

        Uniform
        & $58.99{\pm}0.54$
        & $0.074{\pm}0.004$
        & $44.73{\pm}0.11$
        & $0.033{\pm}0.007$
        & $58.76{\pm}1.07$
        & $0.087{\pm}0.006$
        \\

        iCaRL
        & $60.74{\pm}0.09$
        & $\mathbf{0.044{\pm}0.026}$
        & $44.25{\pm}2.04$
        & $0.042{\pm}0.019$
        & $59.70{\pm}0.70$
        & $0.071{\pm}0.010$
        \\

        Grad Matching
        & $59.17{\pm}0.38$
        & $0.067{\pm}0.003$
        & $45.44{\pm}0.64$
        & $0.038{\pm}0.001$
        & $59.58{\pm}0.28$
        & $0.073{\pm}0.008$
        \\

        SPR
        & $59.56{\pm}0.73$
        & $0.143{\pm}0.064$
        & $44.45{\pm}0.55$
        & $0.086{\pm}0.023$
        & $58.74{\pm}0.63$
        & $0.073{\pm}0.010$
        \\

        MetaSP
        & $60.14{\pm}0.25$
        & $0.056{\pm}0.230$
        & $43.74{\pm}0.36$
        & $0.079{\pm}0.014$
        & $57.43{\pm}0.54$
        & $0.086{\pm}0.007$
        \\

        Greedy Coreset
        & $59.39{\pm}0.16$
        & $0.066{\pm}0.017$
        & $43.80{\pm}0.01$
        & $0.039{\pm}0.007$
        & $58.22{\pm}0.16$
        & $0.066{\pm}0.001$
        \\

        GCR
        & $58.73{\pm}0.43$
        & $0.073{\pm}0.013$
        & $44.48{\pm}0.05$
        & $0.035{\pm}0.005$
        & $58.72{\pm}0.63$
        & $0.081{\pm}0.005$
        \\

        PBCS
        & $55.64{\pm}2.26$
        & $0.062{\pm}0.001$
        & $39.87{\pm}1.12$
        & $0.076{\pm}0.011$
        & $56.93{\pm}0.14$
        & $0.100{\pm}0.003$
        \\

        OCS
        & $52.57{\pm}0.37$
        & $0.088{\pm}0.001$
        & $46.54{\pm}0.34$
        & $\mathbf{0.022{\pm}0.003}$
        & $51.77{\pm}0.81$
        & $0.103{\pm}0.007$
        \\

        BCSR
        & $\mathbf{61.60{\pm}0.14}$
        & $0.051{\pm}0.015$
        & $47.30{\pm}0.57$
        & $\mathbf{0.022{\pm}0.005}$
        & $60.70{\pm}0.08$
        & $\mathbf{0.059{\pm}0.013}$
        \\

        \midrule

        \rowcolor{lilac}
        \textbf{CARES (Ours)}
        & $61.52{\pm}1.48$
        & $0.111{\pm}0.074$
        & $\mathbf{54.88{\pm}0.33}$
        & $0.0822{\pm}0.083$
        & $\mathbf{62.12{\pm}1.14}$
        & $0.110{\pm}0.013$
        \\

        \bottomrule

    \end{tabular}
    }

\end{table}

\begin{table}[t]
\centering
\small

\caption{
Results on Split-Tiny-ImageNet under balanced, imbalanced, and label-noise
settings. Higher average accuracy is better, while lower forgetting
is better. Results are reported as mean $\pm$ standard deviation.
}

\label{tab:tiny_full_results}

\setlength{\tabcolsep}{5pt}

\resizebox{\textwidth}{!}{%

\begin{tabular}{lcccccc}

\toprule

\multirow{2}{*}{\textbf{Method}}
& \multicolumn{2}{c}{\textbf{Balanced}}
& \multicolumn{2}{c}{\textbf{Imbalanced}}
& \multicolumn{2}{c}{\textbf{Label Noise}} \\

\cmidrule(lr){2-3}
\cmidrule(lr){4-5}
\cmidrule(lr){6-7}

&
AVG ACC $\uparrow$
&
FGT $\downarrow$
&
AVG ACC $\uparrow$
&
FGT $\downarrow$
&
AVG ACC $\uparrow$
&
FGT $\downarrow$
\\

\midrule

K-means Embedding
& $41.48{\pm}1.21$
& $0.129{\pm}0.007$
& $36.29{\pm}0.23$
& $0.085{\pm}0.003$
& $36.01{\pm}1.51$
& $0.083{\pm}0.005$
\\

Uniform
& $42.11{\pm}0.52$
& $0.129{\pm}0.002$
& $37.07{\pm}0.53$
& $0.083{\pm}0.009$
& $37.14{\pm}1.05$
& $0.099{\pm}0.003$
\\

iCaRL
& $43.84{\pm}0.09$
& $0.114{\pm}0.004$
& $37.65{\pm}0.84$
& $0.058{\pm}0.003$
& $38.52{\pm}0.25$
& $0.063{\pm}0.006$
\\

Greedy Coreset
& $41.02{\pm}0.33$
& $0.119{\pm}0.017$
& $33.43{\pm}0.86$
& $0.103{\pm}0.002$
& $36.37{\pm}0.16$
& $0.079{\pm}0.006$
\\

GCR
& $41.45{\pm}0.35$
& $0.125{\pm}0.008$
& $36.08{\pm}0.62$
& $0.072{\pm}0.017$
& $37.46{\pm}0.40$
& $0.115{\pm}0.011$
\\

OCS
& $41.29{\pm}0.09$
& $0.112{\pm}0.001$
& $35.09{\pm}0.60$
& $0.036{\pm}0.011$
& $35.36{\pm}0.94$
& $0.061{\pm}0.005$
\\

BCSR
& $44.13{\pm}0.33$
& $0.106{\pm}0.001$
& $38.59{\pm}0.11$
& $0.070{\pm}0.004$
& $40.72{\pm}0.56$
& $0.055{\pm}0.006$
\\

\midrule

\rowcolor{lilac}
\textbf{CARES (Ours)}
& $\mathbf{49.49{\pm}0.41}$
& $\mathbf{0.099{\pm}0.015}$
& $\mathbf{42.94{\pm}0.40}$
& $0.104{\pm}0.011$
& $\mathbf{49.37{\pm}0.95}$
& $0.078{\pm}0.021$
\\

\bottomrule

\end{tabular}
}

\end{table}

\begin{table}[t]
\centering
\caption{Online CIL average accuracy (15 seeds). Best results are shown in bold.}
\label{tab:cil_avg_accuracy_full}
\scriptsize
\setlength{\tabcolsep}{4pt}
\resizebox{\textwidth}{!}{%
\begin{tabular}{lccccccccc}
\toprule
\multirow{2}{*}{\textbf{Method}} 
& \multicolumn{3}{c}{\textbf{S-CIFAR-10}} 
& \multicolumn{3}{c}{\textbf{S-CIFAR-100}} 
& \multicolumn{3}{c}{\textbf{S-TinyImageNet}} \\
\cmidrule(lr){2-4} \cmidrule(lr){5-7} \cmidrule(lr){8-10}
& $K=200$ & $K=500$ & $K=2000$
& $K=200$ & $K=500$ & $K=2000$
& $K=200$ & $K=500$ & $K=2000$ \\
\midrule
ER 
& $32.63{\pm}5.11$ & $43.32{\pm}4.75$ & $49.31{\pm}7.45$
& $11.72{\pm}1.25$ & $14.94{\pm}1.69$ & $20.78{\pm}1.48$
& $4.36{\pm}1.02$ & $6.69{\pm}0.53$ & $11.3{\pm}1.45$ \\

GEM 
& $16.81{\pm}1.17$ & $17.73{\pm}0.98$ & $20.35{\pm}2.44$
& $10.42{\pm}1.66$ & $9.18{\pm}1.8$ & $12.38{\pm}3.81$
& $5.15{\pm}0.71$ & $6.14{\pm}1.01$ & $9.56{\pm}2.63$ \\

GSS 
& $33.83{\pm}3.95$ & $40.59{\pm}4.20$ & $41.7{\pm}4.56$
& $12.31{\pm}0.57$ & $13.84{\pm}0.57$ & $16.24{\pm}0.80$
& $5.82{\pm}0.38$ & $6.78{\pm}0.41$ & $8.44{\pm}0.48$ \\

iCaRL 
& $38.66{\pm}2.03$ & $33.53{\pm}3.17$ & $39.56{\pm}1.53$
& $11.23{\pm}0.31$ & $11.81{\pm}0.29$ & $12.28{\pm}0.39$
& $4.39{\pm}0.19$ & $5.36{\pm}0.27$ & $6.66{\pm}0.30$ \\

MIR 
& $21.5{\pm}0.7$ & $30.52{\pm}1.22$ & $46.4{\pm}1.81$
& $10.3{\pm}0.3$ & $11.4{\pm}0.38$ & $16.02{\pm}0.5$
& $4.9{\pm}0.17$ & $5.1{\pm}0.33$ & $7.05{\pm}0.30$ \\

DER 
& $38.26{\pm}5.08$ & $45.56{\pm}6.02$ & $49.9{\pm}7.55$
& $10.68{\pm}2.03$ & $14.43{\pm}2.44$ & $24.83{\pm}1.58$
& $4.81{\pm}0.57$ & $8.49{\pm}0.48$ & $\mathbf{15.97{\pm}0.95}$ \\

GCR 
& $\mathbf{42.26{\pm}7.6}$ & $\mathbf{51.17{\pm}3.5}$ & $52.3{\pm}7.01$
& $\mathbf{14.88{\pm}0.94}$ & $18.89{\pm}1.5$ & $22.47{\pm}2.01$
& $6.19{\pm}0.36$ & $\mathbf{8.77{\pm}0.5}$ & $11.83{\pm}1.26$ \\

\midrule
\rowcolor{lilac}
\textbf{CARES (ours)}
& $40.26{\pm}4.67$ & $48.49{\pm}5.83$ & $\mathbf{54.88{\pm}6.34}$
& $14.28{\pm}0.84$ & $\mathbf{19.12{\pm}1.28}$ & $\mathbf{25.17{\pm}1.39}$
& $\mathbf{6.60{\pm}0.59}$ & $8.46{\pm}0.78$ & $15.67{\pm}0.85$ \\
\bottomrule
\end{tabular}%
}
\end{table}

\section{Implementation Details}
\label{sec:supp_implementation}
Tables~\ref{tab:protocol1_hparams} and \ref{tab:protocol2_hparams} show the hyperparameter values of the best results under each protocol. All experiments were run on a single NVIDIA RTX A6000 GPU with four
CPU cores. Unless otherwise stated, the same hardware configuration
was used across all methods and experimental settings.

\begin{table}[!h]
\centering
\caption{Hyperparameter settings for Protocol~1. Settings shared across
stream conditions are reported once per dataset; learning rate, weight
decay, coverage, and staging-pool size are selected separately for the
balanced, imbalanced, and label-noise settings.}
\label{tab:protocol1_hparams}

\small
\setlength{\tabcolsep}{4.5pt}
\renewcommand{\arraystretch}{1.08}

\begin{tabular}{lcccccc}
\toprule
\multirow{2}{*}{\textbf{Hyperparameter}}
& \multicolumn{3}{c}{\textbf{Split CIFAR-100}}
& \multicolumn{3}{c}{\textbf{Split Tiny-ImageNet}} \\
\cmidrule(lr){2-4}
\cmidrule(lr){5-7}
& \textbf{Bal.}
& \textbf{Imb.}
& \textbf{Noise}
& \textbf{Bal.}
& \textbf{Imb.}
& \textbf{Noise} \\
\midrule

Number of tasks
& \multicolumn{3}{c}{20}
& \multicolumn{3}{c}{20} \\

Classes per task
& \multicolumn{3}{c}{5}
& \multicolumn{3}{c}{10} \\

Replay capacity $K$
& \multicolumn{3}{c}{100}
& \multicolumn{3}{c}{200} \\

Incoming stream batch size
& \multicolumn{3}{c}{50}
& \multicolumn{3}{c}{100} \\

Training batch size
& \multicolumn{3}{c}{\textit{10}}
& \multicolumn{3}{c}{\textit{20}} \\

Epochs per stream batch
& \multicolumn{3}{c}{1}
& \multicolumn{3}{c}{1} \\

Backbone
& \multicolumn{3}{c}{ResNet-18}
& \multicolumn{3}{c}{ResNet-18} \\

Number of runs
& \multicolumn{3}{c}{3}
& \multicolumn{3}{c}{3} \\

\midrule

Learning rate
& \textit{0.03}
& \textit{0.03}
& \textit{0.05}
& \textit{0.03}
& \textit{0.03}
& \textit{0.05} \\

Weight decay
& \textit{1e-5}
& \textit{1e-5}
& \textit{1e-5}
& \textit{1e-5}
& \textit{5e-5}
& \textit{5e-5} \\

Coverage $\gamma$
& \textit{0.4}
& \textit{0.4}
& \textit{0.4}
& \textit{0.2}
& \textit{0.4}
& \textit{0.2} \\

Staging-pool size $P$
& \textit{200}
& \textit{200}
& \textit{200}
& \textit{400}
& \textit{400}
& \textit{400} \\

\bottomrule
\end{tabular}
\end{table}

\begin{table}[h]
\centering
\caption{Hyperparameter settings for Protocol~2.}
\label{tab:protocol2_hparams}
\small
\setlength{\tabcolsep}{5.5pt}
\renewcommand{\arraystretch}{1.08}

\begin{tabular}{lccc}
\toprule
\textbf{Hyperparameter}
& \textbf{Split CIFAR-10}
& \textbf{Split CIFAR-100}
& \textbf{Split Tiny-ImageNet} \\
\midrule

Number of tasks
& 5
& 5
& 10 \\

Replay capacity $K$
& \{200, 500, 2000\}
& \{200, 500, 2000\}
& \{200, 500, 2000\} \\

Training batch size
& 32
& 32
& 32 \\

Incoming stream batch size
& 6250
& 6250
& 6250 \\

Training epochs
& 1
& 1
& 1 \\

Learning rate
& 0.05
& 0.05
& 0.05 \\

Optimizer
& SGD
& SGD
& SGD \\

Weight decay
& 1e-5
& 1e-5
& 1e-5 \\

Coverage $\gamma$
& 0.4
& 0.4
& 0.4 \\

Staging-pool size $P$
& 400
& 400
& 400 \\

Backbone
& ResNet-18
& ResNet-18
& ResNet-18 \\

Number of runs
& 15
& 15
& 15 \\

\bottomrule
\end{tabular}
\end{table}

\section{Sensitivity Analysis}
\label{sec:supp_sensitivity}

\subsection{Replay and Staging-Pool Scaling}

We additionally examine how replay capacity $K$ and staging-pool
capacity $P$ affect CARES on Split CIFAR-100 and Split-Tiny-ImageNet under Protocol 1. Increasing replay
capacity substantially improves performance: at a fixed staging-pool
size of $P=200$, increasing $K$ from 200 to 1000 raises final CIL
accuracy from 9.83\% to 19.88\% and TIL accuracy from 63.60\% to
73.63\%. In contrast, increasing the staging-pool size at a fixed
replay budget provides comparatively modest gains. For $K=500$,
increasing $P$ from 200 to 1000 improves CIL accuracy from 16.06\% to
16.72\% and TIL accuracy from 69.26\% to 70.12\%, while increasing
the memory-update overhead from 11.2\,s to 23.5\,s. These results
suggest that CARES benefits primarily from additional persistent replay
capacity, while very large staging pools offer diminishing returns.

\begin{table}[h]
\centering
\caption{
Scaling of CARES with replay capacity $K$ and staging-pool capacity $P$
on Split CIFAR-100 under Protocol 1. We report final class-incremental (CIL) and
task-incremental (TIL) accuracy and the cumulative memory-update
overhead. All other hyperparameters are held fixed. Results are
reported as mean $\pm$ standard deviation over $n$ completed runs.
}
\label{tab:cares_kp_scaling}

\small
\setlength{\tabcolsep}{5pt}
\renewcommand{\arraystretch}{1.08}

\begin{tabular}{cccrrr}
\toprule
\textbf{$K$}
& \textbf{$P$}
& \textbf{CIL Acc. $\uparrow$}
& \textbf{TIL Acc. $\uparrow$}
& \textbf{Update Overhead (s) $\downarrow$} \\
\midrule

\multirow{3}{*}{200}
& 200 & $9.83{\pm}0.42$  & $63.60{\pm}0.70$ & $6.8{\pm}1.0$ \\
& 500  & $9.78{\pm}0.63$  & $64.16{\pm}1.90$ & $12.4{\pm}2.1$ \\
& 1000 & $10.47{\pm}0.00$ & $62.46{\pm}0.00$ & $17.6{\pm}0.0$ \\
\midrule

\multirow{3}{*}{500}
& 200 & $16.06{\pm}0.90$ & $69.26{\pm}0.56$ & $11.2{\pm}1.9$ \\
& 500 & $16.24{\pm}1.33$ & $69.37{\pm}1.02$ & $15.1{\pm}7.3$ \\
& 1000 & $16.72{\pm}1.50$ & $70.12{\pm}0.97$ & $23.5{\pm}13.4$ \\
\midrule

\multirow{3}{*}{1000}
& 200 & $19.88{\pm}0.14$ & $73.63{\pm}0.85$ & $14.7{\pm}3.1$ \\
& 500 & $20.08{\pm}0.00$ & $73.90{\pm}0.11$ & $23.2{\pm}7.2$ \\
& 1000 & $20.05{\pm}0.00$ & $73.92{\pm}0.00$ & $44.8{\pm}0.0$ \\

\bottomrule
\end{tabular}
\end{table}

\subsection{Joint Sensitivity to Coverage and Staging-Pool Capacity}
\label{sec:supp_additional_sensitivity}

We further study the interaction between the coverage parameter
$\gamma$ and staging-pool capacity $P$ on balanced Split CIFAR-100.
We fix the replay capacity to $K=100$, the learning rate to $0.02$,
and weight decay to $5\times10^{-5}$, while jointly varying
$\gamma\in\{0.2,0.3,0.4,0.6,0.8\}$ and
$P\in\{K,2K,5K,10K,20K\}$.

\begin{figure}[!htbp]
    \centering
    \includegraphics[width=0.55\linewidth]{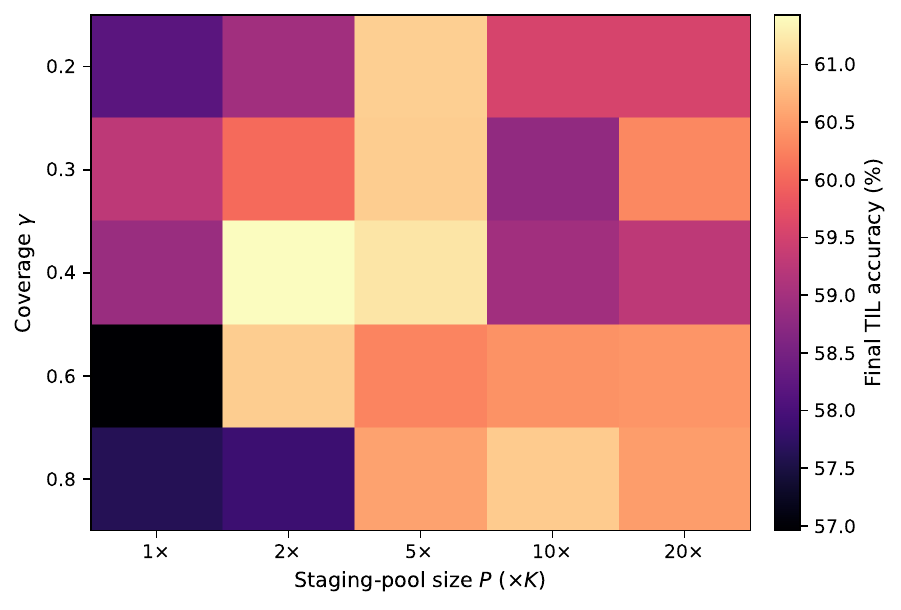}
    \caption{
    Joint sensitivity of CARES to the coverage parameter $\gamma$ and
    staging-pool capacity $P$ on balanced Split CIFAR-100. Each cell
    represents final task-incremental accuracy, with $P$ expressed relative
    to replay capacity $K=100$. CARES exhibits a broad region of strong
    performance, while increasing the staging pool beyond moderate sizes
    provides no consistent benefit.
    }
    \label{fig:coverage_pool_heatmap}
\end{figure}

Figure~\ref{fig:coverage_pool_heatmap} shows that strong performance is
obtained over a relatively broad region of the parameter space rather
than at a single narrow operating point. In particular, staging pools
of $2K$--$5K$ consistently achieve high final TIL accuracy across
multiple coverage values. The highest mean accuracy is obtained at
$\gamma=0.4$ and $P=2K$, but several neighboring configurations
perform similarly. Increasing the staging pool further to $10K$ or
$20K$ does not provide a consistent accuracy improvement. These results suggest that CARES does not require precise tuning of
either $\gamma$ or $P$, and that a moderately sized staging pool is
sufficient to obtain strong performance.

\end{document}